\documentclass{article} 
\usepackage{iclr2022_conference,times}

\usepackage{amsmath,amsfonts,bm}

\def\eqref#1{equation~\ref{#1}}

\def\1{\bm{1}}

\def\va{{\bm{a}}}
\def\vb{{\bm{b}}}
\def\vc{{\bm{c}}}
\def\vd{{\bm{d}}}

\def\vh{{\bm{h}}}

\def\vm{{\bm{m}}}

\def\vp{{\bm{p}}}

\def\vr{{\bm{r}}}
\def\vs{{\bm{s}}}

\def\vx{{\bm{x}}}

\def\vz{{\bm{z}}}

\def\mU{{\bm{U}}}

\def\mW{{\bm{W}}}

\DeclareMathAlphabet{\mathsfit}{\encodingdefault}{\sfdefault}{m}{sl}
\SetMathAlphabet{\mathsfit}{bold}{\encodingdefault}{\sfdefault}{bx}{n}

\def\gB{{\mathcal{B}}}

\def\gE{{\mathcal{E}}}

\def\gG{{\mathcal{G}}}

\def\gS{{\mathcal{S}}}

\def\gV{{\mathcal{V}}}

\def\gZ{{\mathcal{Z}}}

\newcommand{\R}{\mathbb{R}}

\usepackage{hyperref}
\usepackage{url}
\usepackage{makecell}
\usepackage{tabularx,booktabs}

\usepackage[T1]{fontenc}

\usepackage{pifont}
\newcommand{\cmark}{\ding{51}}
\newcommand{\xmark}{\ding{55}}

\usepackage[pdftex]{graphicx}

\usepackage{authblk}

\title{Permutation invariant graph-to-sequence model for template-free retrosynthesis and reaction prediction}

\author[1,2]{Zhengkai Tu}
\author[1,3]{Connor W. Coley}
\affil[1]{Department of Chemical Engineering, MIT}
\affil[2]{Center for Computational Science \& Engineering, MIT}
\affil[3]{Department of Electrical Engineering and Computer Science, MIT}
\affil[ ]{\texttt{\{ztu,ccoley\}@mit.edu}}

\iclrfinalcopy 
\begin{document}

\maketitle

\begin{abstract}
Synthesis planning and reaction outcome prediction are two fundamental problems in computer-aided organic chemistry for which a variety of data-driven approaches have emerged. Natural language approaches that model each problem as a SMILES-to-SMILES translation lead to a simple end-to-end formulation, reduce the need for data preprocessing, and enable the use of well-optimized machine translation model architectures. However, SMILES representations are not an efficient representation for capturing information about molecular structures, as evidenced by the success of SMILES augmentation to boost empirical performance. Here, we describe a novel Graph2SMILES model that combines the power of Transformer models for text generation with the permutation invariance of molecular graph encoders that mitigates the need for input data augmentation. As an end-to-end architecture, Graph2SMILES can be used as a drop-in replacement for the Transformer in any task involving molecule(s)-to-molecule(s) transformations. In our encoder, an attention-augmented directed message passing neural network (D-MPNN) captures local chemical environments, and the global attention encoder allows for long-range and intermolecular interactions, enhanced by graph-aware positional embedding.  Graph2SMILES improves the top-1 accuracy of the Transformer baselines by $1.7\%$ and $1.9\%$ for reaction outcome prediction on USPTO\_480k and USPTO\_STEREO datasets respectively, and by $9.8\%$ for one-step retrosynthesis on the USPTO\_50k dataset.

\end{abstract}

\section{Introduction}
Retrosynthetic analysis \citep{Corey_1988_RetroReview,Corey_1989_logic} and its reverse problem, reaction outcome prediction \citep{Corey_Wipke_1969_CAD}, are two fundamental problems in computer-aided organic synthesis. The former tries to propose possible reaction precursors given the desirable product, whereas the latter aims to predict the major products given reactants. Historically, they were tackled using rule-based expert systems such as LHASA \citep{Corey_1972_LHASA}. Recent developments in machine learning have led to a number of new template-based, graph edit-based, and translation-based methods, for which we give a detailed review in Section \ref{sec:casp}. For both tasks, translation-based approaches have grown popular, possibly because the end-to-end formulation is procedurally simple. Since most organic molecules can be represented as SMILES strings \citep{Neglur_2005_badSMILES}, retrosynthesis can be cast as a translation from product SMILES to reactant SMILES \citep{Liu_2017_S2S}, and so can reaction outcome prediction \citep{Nam_Kim_2016_S2S}. Modeling these tasks as machine translation problems enables the use of neural architectures that are well-studied and well-optimized in the field of Natural Language Processing (NLP).  Several of the best performing models across multiple benchmark datasets \citep{Tetko_2020_AT,Irwin_2021_Chemformer,Sun_2020_EBM,Seo_2021_GTA,Wang_2021_RetroPRIME} have used the Transformer architecture \citep{Vaswani_2017_Transformer} as the backbone on SMILES representations, showing the effectiveness of translation-based formulation.

However, SMILES representations do not provide bijective mappings to molecular structure. 
As a result, data augmentation with chemically-equivalent SMILES \citep{Bjerrum_2017_Aug} has become a common practice to improve empirical performance. Augmenting the training data with just 1 equivalent reaction SMILES by permuting both the inputs and the outputs can already produce noticeable improvements of $0.8\%$ to $4.3\%$ \citep{Schwaller_2019_MT,Seo_2021_GTA}. Incorporating 9 \citep{Wang_2021_RetroPRIME}, or up to 100 \citep{Tetko_2020_AT} equivalent SMILES can provide additional gains. While these efforts demonstrate the effectiveness of SMILES augmentation, they can also be interpreted as evidence of the ineffectiveness of the SMILES representation itself. 

In this paper, we propose a novel graph-to-sequence architecture called Graph2SMILES to solve the tasks of retrosynthesis and reaction prediction. We first design a sequential graph encoder with an attention-augmented directed message passing neural network (D-MPNN) based on \citet{Yang_2019_Chemprop}, followed by a Transformer-based global attention encoder with graph-aware positional embedding. We then pair the graph encoder with a Transformer decoder to transform molecular graph inputs into SMILES outputs, without using sequence representations of input SMILES at all. As such, we guarantee the permutation invariance of Graph2SMILES to the input, eliminating the need for input-side augmentation altogether. Our main contributions can be summarized as follows:
\begin{enumerate}
    \item We propose a Graph2SMILES architecture with two encoder components modeling local and global atomic interactions respectively, to address forward prediction and one-step retrosynthesis as graph-to-sequence tasks.
    \item We design a graph-aware positional embedding to further enhance performance. It is easily generalizable to graphs containing two or more molecules, while not requiring any pretraining or joint training with auxiliary tasks.
    \item We demonstrate the adequacy of graph representations alone by showing that Graph2SMILES outperforms Transformer baselines on predictive chemistry tasks without needing any input-side SMILES augmentation.
\end{enumerate}
Our Graph2SMILES architecture achieves state-of-the-art top-1 accuracy on common benchmarks among methods that do not make use of reaction templates, atom mapping, pretraining, or data augmentation strategies. We emphasize that Graph2SMILES is a backbone architecture, and hence a drop-in replacement for the Transformer model. As such, Graph2SMILES can be plugged into any method for molecular transformation tasks that uses the Transformer model, while retraining the benefits from techniques or features orthogonal to the architecture itself.

\section{Methods}
\subsection{Graph and sequence representations of molecules}
There are multiple ways of representing molecular structures, such as by molecular fingerprints \citep{Rogers_and_Hahn_2010_ECFP}, by SMILES strings \citep{Weininger_1988_SMILES}, or as molecular graphs with atoms as nodes and bonds as edges. We represent the input molecule(s) as graphs, and the output molecule(s) as SMILES strings, thus modeling both reaction prediction and retrosynthesis as graph-to-sequence transformations.

Formally, let $\gG_\text{in}$ denotes the molecular graph input, which can contain multiple subgraphs for different molecules, with a total of $N$ atoms. Following the convention in \citet{Somnath_2020_GRAPHRETRO}, we describe $\gG_\text{in}=(\gV, \gE)$ with atoms $\gV$ and bonds $\gE$. Each atom $u \in \gV$ has a feature vector $\vx_u \in \R^{a}$, and each directed bond $(u, v) \in \gE$ from atom $u$ to $v$ has its feature vector as $\vx_{uv} \in \R^{b}$. The details of the atom and bond features used can be found in Appendix \ref{appendix:feature}. We build the input molecular graphs from their SMILES strings with RDKit \citep{Landrum_2016_RDKit}. Note that all feature vectors are invariant to the order of atoms and bonds, as well as to how the original SMILES strings are written. We represent the output as a sequence of SMILES tokens $\gS_\text{out}$ = $\{s_1, s_2, \dots, s_n\}$, where the tokens $\{s_i\}$ are obtained from the canonical SMILES using the regex tokenizer in \citet{Schwaller_2019_MT}.

\subsection{Graph2SMILES}
\begin{figure}[t]
\begin{center}
\includegraphics[width=1.0\linewidth]{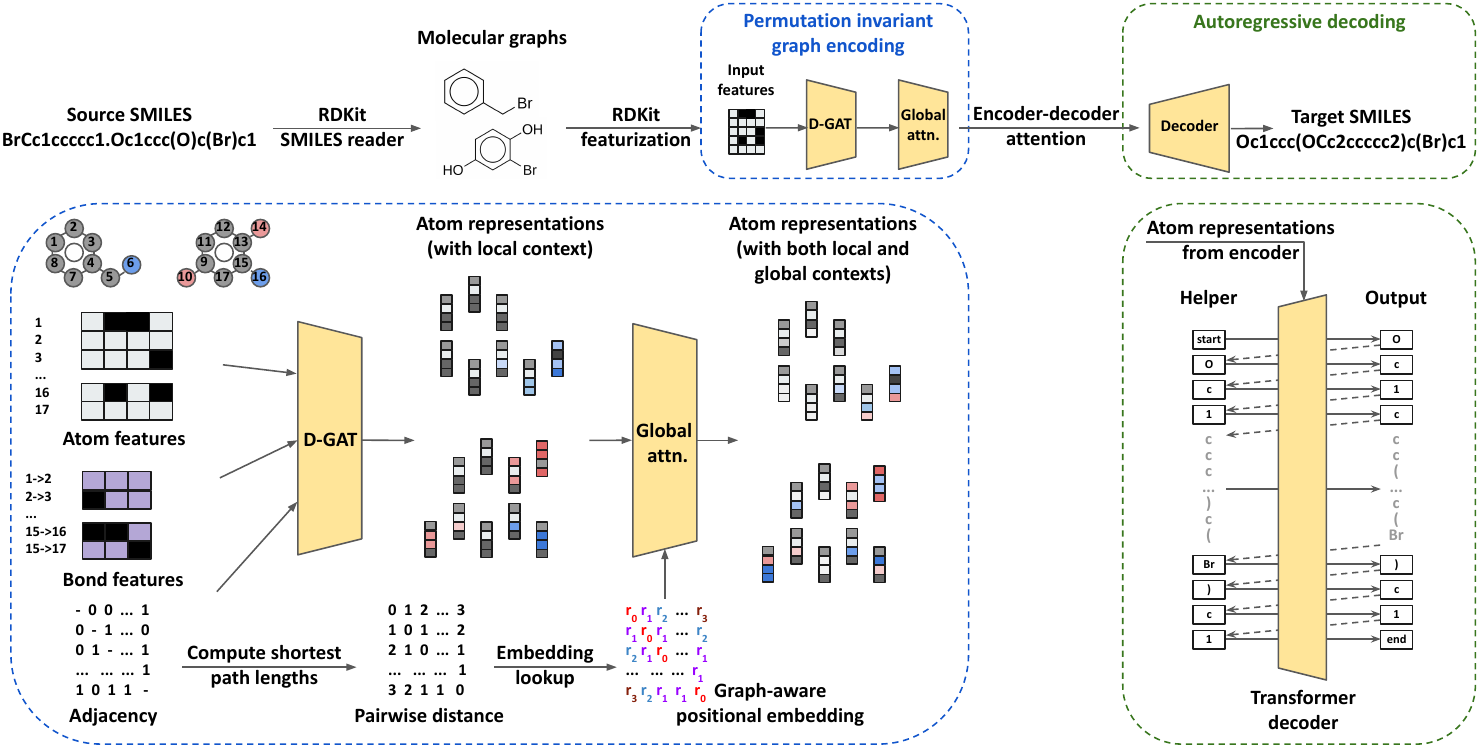}
\end{center}
\caption{Model architecture for Graph2SMILES. Top: the overall flowchart. Bottom left: details of permutation invariant graph encoding. Bottom right: details of autoregressive decoding.}
\label{fig:architecture}
\end{figure}

The Graph2SMILES model is a variant of the encoder-decoder model \citep{Cho_2014_EncDec} commonly used for machine translation. Figure \ref{fig:architecture} displays the architecture of Graph2SMILES with the permutation invariant graph encoding process shown within the blue dashed box. We replace the encoder part of the standard Transformer model \citep{Vaswani_2017_Transformer} used in Molecular Transformer \citep{Schwaller_2019_MT} with a novel attention-augmented directed message passing encoder, followed by a global attention encoder with carefully designed graph-aware positional embedding. Each module has its intuitive function: the D-MPNN captures the local chemical context, the global attention encoder allows for global-level information exchange, and the graph-aware positional embedding enables the attention encoder to make use of topological information more explicitly. The permutation invariant encoding process eliminates the need for SMILES augmentation for the input side altogether, simplifying data preprocessing and potentially saving training time.

\subsubsection{Attention augmented directed message passing encoder}
The first module of the graph encoder is a D-MPNN \citep{Yang_2019_Chemprop} with Gated Recurrent Units (GRUs) \citep{Cho_2014_EncDec} used for message updates \citep{Jin_2018_JTVAE,Somnath_2020_GRAPHRETRO}. Unlike atom-oriented message updates in edge-aware MPNNs \citep{Hu_2020_PretrainingGNN,Yan_2020_RetroXpert,Mao_2021_GET,Wang_2021_Meta}, updates in D-MPNN are oriented towards directed bonds to prevent totters, or messages being passed back-and-forth between neighbors \citep{Mahe_2004_Totters,Yang_2019_Chemprop}. We augment the D-MPNN with attention-based message updates inspired from Graph Attention Network \citep{Velickovic_2018_GAT,Brody_2021_GATv2}. We term this variant as Directed Graph Attention Network (D-GAT) and refer to the original D-MPNN variant used in \citet{Somnath_2020_GRAPHRETRO} as Directed Graph Convolutional Network (D-GCN).

For D-GAT, at each message passing step $t$, the message $\vm_{uv}^{t+1}$ associated with each directed bond $(u, v) \in \gE$ is updated using
\begin{align}
    \vs_{uv} &= \mathrm{AttnSum} \left(
    \vx_u, \vx_{uv},
    \left\{\vm_{wu}^{t} \right\}_{w \in N(u) \backslash v}
    \right) \\
    \vz_{uv} &= \sigma \left(
    \mW_{z} \left[
    \vx_u; \vx_{uv}; \vs_{uv}
    \right] + \vb_{z}
    \right) \\
    \vr_{uv} &= \sigma \left(
    \mW_{r} \left[
    \vx_u; \vx_{uv}; \vs_{uv}
    \right] + \vb_{r}
    \right) \label{eqn:reset} \\
    \tilde{\vm}_{uv} &= \mathrm{tanh} \left(
    \mW \left[
    \vx_u; \vx_{uv}
    \right] + \mU \vr_{uv} + \vb
    \right) \\
    \vm_{uv}^{t+1} &= \left(
    1 - \vz_{uv} \right) \odot \vs_{uv} +
    \vz_{uv} \odot \tilde{\vm}_{uv}
\end{align}
where $\mW_z$, $\mW_r$, $ \mW$, $\mU$ and $\vb_z$, $\vb_r$, $\vb$ are the learnable weights and biases respectively. $\sigma$ is the sigmoid function, ";" indicates concatenation, and $\odot$ is the element-wise product. $\mathrm{AttnSum}$ is the attention-based message aggregation defined as

\begin{align}
    e_{wu} &=
    \va^{T} \mathrm{LeakyReLU} \left(
    \mW_{qk} \left[ \vx_u; \vx_{uv}; \vm_{wu}^t \right] + \vb_{qk}
    \right) \\
    a_{wu} &=
    \frac{\mathrm{exp} \left( e_{wu} \right)}
    {\sum_{w \in N(u) \backslash v} \mathrm{exp} \left( e_{wu} \right)} \\
    \vs_{uv} &= \sum_{w \in N(u) \backslash v}
    a_{wu}
    \left( \mW_{v} \vm_{wu}^t + \vb_{v} \right)
\end{align}

where $\mW_{qk}$, $\vb_{qk}$, $\va$ are the learnable parameters for the attention scores, and $\mW_{v}$, $\vb_{v}$ are the parameters for the value vectors. We have omitted a self-loop in message aggregation, as it did not have any noticeable effect on performance from our experiments. In Eqn (\ref{eqn:reset}) we simplify the reset gate to be shared for all incoming edges, in contrast to \citet{Somnath_2020_GRAPHRETRO} where a separate reset gate is defined for each edge.

Lastly, after $T$ iterations, we obtain the atom representations $\vh_u$ with similar attention-based aggregation (with different parameters) over the bond messages coming into each atom $u$, followed by a single output layer with weight $\mW_o$ and GELU activation \citep{Hendrycks_Gimpel_2016_GELU}.
\begin{align}
    \vm_{u} &= \mathrm{AttnSum'} \left(
    \vx_{u}, \left\{\vm_{wu}^{(T)} \right\}_{w \in N(u)}
    \right) \\
    \vh_{u} &= \mathrm{GELU} \left(
    \mW_{o} \left[\vx_{u}; \vm_{u} \right]
    \right)
\end{align}
We use multi-headed attention in our formulation similar to \citet{Brody_2021_GATv2}.

\subsubsection{Global attention encoder with graph-aware positional embedding}
To capture global interactions, the atom representations coming out of the D-MPNN are fed into a global attention encoder, which is a variant of the Transformer encoder. We incorporate graph-aware positional embedding, adapted from the relative positional embedding used in Transformer-XL \citep{Dai_2019_TransformerXL} as follows. Firstly, in the standard Transformer either with sinusoidal encoding \citep{Vaswani_2017_Transformer,Schwaller_2019_MT} or learnable \citep{Devlin_2019_BERT} absolute positional embedding, the attention score between atoms $u$ and $v$ can be decomposed as
\begin{align}
    e_{u,v}^{abs}
    &= \vh_{u}^{T} \tilde{\mW}_{q}^{T}
    \tilde{\mW}_{k} \vh_{v}
    + \vh_{u}^{T} \tilde{\mW}_{q}^{T}
    \tilde{\mW}_{k} \vp_{v}
    + \vp_{u}^{T} \tilde{\mW}_{q}^{T}
    \tilde{\mW}_{k} \vh_{v}
    + \vp_{u}^{T} \tilde{\mW}_{q}^{T}
    \tilde{\mW}_{k} \vp_{v}
\end{align}
where $\tilde{\mW}_{q}$ ,$\tilde{\mW}_{k}$ are weights for the keys and queries, and $\vp_{u}$, $\vp_{v}$ are the absolute positional encoding or embedding corresponding to atoms $u$ and $v$. Similar to Transformer-XL, we reparameterize the four terms. Instead of using sequence-based relative positional embedding $\vr_{u-v}$, we use a learnable embedding term $\vr_{u,v}$ that is dependent on the shortest path length between $u$ and $v$
\begin{align}
    e_{u,v}^{rel}
    &= \left(\vh_{u}^{T} \tilde{\mW}_{q}^{T}
    + \vc^{T} \right)
    \tilde{\mW}_{k} \vh_{v}
    + \left( \vh_{u}^{T} \tilde{\mW}_{q}^{T}
    + \vd^{T} \right)
    \tilde{\mW}_{k,R} \vr_{u,v}
    \nonumber \\
    &= \left(\vh_{u}^{T} \tilde{\mW}_{q}^{T}
    + \vc^{T} \right)
    \tilde{\mW}_{k} \vh_{v}
    + \left( \vh_{u}^{T} \tilde{\mW}_{q}^{T}
    + \vd^{T} \right)
    \tilde{\vr}_{u,v} \label{eqn:interaction}
\end{align}

The trainable biases are renamed as $\vc$ and $\vd$ to avoid confusion and shared across all layers. Intuitively, the two terms in Eqn (\ref{eqn:interaction}) model the interactions between inputs ($\vh_{u}$ and $\vh_{v}$), and between input and relative graph position ($\vh_{u}$ and $\tilde{\vr}_{u,v}$) respectively. Unlike Transformer-XL, we forgo the inductive bias built into the sinusoidal encoding, merging $\tilde{\mW}_{k,R} \vr_{u,v}$ into a single learnable $\tilde{\vr}_{u,v}$. This makes the relative positional embedding easily generalizable to atoms not within the same molecule, which is particular useful for reaction outcome prediction and, more generally, tasks with more than two input molecules. We also bucket the distances similar to \citet{Raffel_2020_T5} such that
\begin{align}
    \gB_{u,v} =
    \begin{cases}
    \mathrm{distance(u,v)}, \, & \mathrm{if \, distance(u,v) < 8} \\
    8, \, & \mathrm{if \, 8 \leq distance(u,v) < 15} \\
    9, \, & \mathrm{if \, 15 \leq distance(u,v) \, and \, u,v \, are \, in \, the \, same \, molecule} \\
    10, \, & \mathrm{if \, u,v \, are \, not \, in \, the \, same \, molecule} \nonumber
    \end{cases}
\end{align}

$\gB_{u,v}$ is then used to look up $\tilde{\vr}_{u,v}$ in the trainable positional embedding matrix. The rest of the global attention encoder mostly follows a standard Transformer with multi-headed self-attention \citep{Vaswani_2017_Transformer}, layer normalization \citep{Ba_2016_LayerNorm}, and position-wise feed forward layers, except no positional information is added to the value vectors within each Transformer layer.

\subsubsection{Sequence decoder}
We use a Transformer-based autoregressive decoder to decode from the atom representations after the global attention encoder. Each output token is generated by attending to all atoms with encoder-decoder attention \citep{Bahdanau_2015_attention,Vaswani_2017_Transformer}, while also attending to all tokens that have already been generated. Following \citet{Seo_2021_GTA}, we set max relative positions to 4 for the decoder, thereby enabling the usage of sequence-based relative positional embedding used in \citet{Shaw_2018_RelPE} and implemented by OpenNMT \citep{Klein_2017_OpenNMT}.

\subsection{Model training}
The Graph2SMILES model is then trained to maximize the conditional likelihood
\begin{align}
    p\left(\gS_\text{out} | \gG_\text{in} \right)
    &= p\left(s_1, s_2, \dots, s_n | \gG_\text{in} \right)
    = \prod_{i=1}^{n} p_{\theta} \left(
    s_i | s_{1:i-1}, \gG_\text{in}
    \right)
\end{align}

\section{Related work}

\subsection{Reaction outcome prediction and one-step retrosynthesis} \label{sec:casp}
One approach to computer-aided reaction prediction and retrosynthesis is to make use of chemical reaction rules based on subgraph pattern matching that are formalized as reaction templates, as in expert systems such as LHASA \citep{Corey_1972_LHASA} and SYNTHIA \citep{Szymkuc_2016_SYNTHIA}. More recent efforts have used neural networks to model the two tasks as template classification \citep{Segler_Waller_2017_NeuralSym,Baylon_2019_Multiscale,Dai_2019_GLN,Chen_Jung_2021_LocalRetro}, or template ranking based on molecular similarity \citep{Coley_2017_RetroSim}. These template-based approaches select the top ranked templates, which can then be applied to transform the input molecules into the outputs.

For template-based methods, there is an inevitable tradeoff between template generality and specificity. Further,  these methods cannot generalize to unseen templates. As a remediation for such intrinsic limitations, quite a number of template-free approaches have emerged over the recent years, which can be broadly categorized into graph edit-based and translation-based. The first category models reaction prediction and/or retrosynthesis as graph transformations \citep{Jin_2017_WLDN,Coley_2018_WLDN5,Do_2019_GTPN,Bradshaw_2018_ELECTRO,Sacha_2021_MEGAN,Qian_2020_Symbolic}. Variants of graph edit methods include electron flow prediction \citep{Bi_2021_NERF} and semi template-based methods where reaction centers are first identified, followed by a graph or sequence recovery stage \citep{Shi_2020_G2Gs,Yan_2020_RetroXpert,Somnath_2020_GRAPHRETRO,Wang_2021_RetroPRIME}. Translation-based formulations, on the other hand, approach the problems as SMILES-to-SMILES translation, typically with sequence models such as Recurrent Neural Networks \citep{Nam_Kim_2016_S2S,Schwaller_2018_S2S,Liu_2017_S2S} or the Transformer \citep{Schwaller_2019_MT,Lin_2020_AutoSynRoute,Lee_2019_MT,Duan_2020_Transformer,Tetko_2020_AT}. Variants of these approaches design additional stages such as pretraining and reranking \citep{Irwin_2021_Chemformer,Zheng_2020_SCROP,Sun_2020_EBM}, or use information about graph topology to enhance performance \citep{Yoo_2020_GRAT,Seo_2021_GTA,Mao_2021_GET}.

Our approach is similar to GET \citep{Mao_2021_GET} which also solves one-step retrosynthesis with graph-enhanced encoders and sequence decoders. Unlike GET which concatenates the SMILES sequence embeddings and learned atom representations, thereby not guaranteeing permutation invariance, we do not use the sequence representation at all in our encoder. Yet Graph2SMILES yields significant improvement over GET on USPTO\_50k, demonstrating the power of our graph encoder and the adequacy of graph representations alone.

\subsection{Adapting the Transformer encoder for molecular representation}
The idea of modeling graphs using the Transformer architecture is not new; the Transformer inherently treats the input tokens as a fully connected graph. In the domain of molecular representation, the Molecular Attention Transformer \citep{Maziarka_2020_MAT} and GeoT \citep{Kwak_2021_GeoT} inject atomic distance information into the computation of attention scores. However, the computation of such distance information itself requires sampling of 3D conformers, thereby introducing an additional source of variations and breaking order invariance. An alternative is using the lengths of shortest paths between atoms. GRAT \citep{Yoo_2020_GRAT} uses these lengths to parameterize the optional scale and bias for the attention score, whereas PAGTN \citep{Chen_2019_PAGTN} treats them as path features in its additive attention. Our use of pairwise shortest path lengths between atoms for our Transformer-based global attention encoder are inspired by GRAT and PAGTN. Contrary to their usage of these lengths as additional features, we explicitly separate the effect of graph topology using graph-aware relative positional embedding, considering the success of such specially designed embedding in graph representation \citep{Ying_2021_Graphormer} and other domains \citep{Shaw_2018_RelPE,Dai_2019_TransformerXL,Wang_2019_StructuralPE,Guo_2020_PCT}. Our formulation of relative positional embedding builds on top of its counterpart in Transformer-XL \citep{Dai_2019_TransformerXL}, which we found to be empirically superior than using single learnable bias as in T5 \citep{Raffel_2020_T5} and Graphormer \citep{Ying_2021_Graphormer}.

\subsection{Combination of graph neural networks and Transformer}
Combining graph encoder and Transformer encoder in a sequential manner has been explored in GET \citep{Mao_2021_GET}, as well as in \citet{Wang_2021_Meta} and GROVER \citep{Rong_2020_GROVER} albeit for different molecular learning tasks. None of these retain the explicit information about graph topology before passing the atom representations into the attention encoder like we do, which we show to be important in the ablation study in Section \ref{subsection:ablation}. Similarly, the graph-to-sequence formulation itself has been used in NLP for conditional text generation tasks such as SQL-to-text \citep{Xu_2018_SQL2Txt,Xu_2018_Graph2Seq} and AMR-to-text \citep{Cai_Lam_2020_Graph2Seq}. Our encoder is different from these prior studies; most notably, the graph-aware positional embedding is designed to easily generalize to more than two disconnected graphs, which is typical for reaction outcome prediction.

\section{Experiments}

\subsection{Datasets}
We evaluate model performance by top-n test accuracies on four USPTO datasets derived from reaction data originally curated by \citet{Lowe_2012_USPTO}. The details of these datasets are summarized in Appendix \ref{appendix:dataset}. For reaction outcome prediction, we evaluate on the USPTO\_480k\_mixed and USPTO\_STEREO\_mixed datasets following \citet{Schwaller_2019_MT}. The suffix \_mixed indicates that the reactants and reagents have not been separated based on which species contribute heavy atoms to the product. While USPTO\_480k has been preprocessed by \citet{Jin_2017_WLDN}, USPTO\_STEREO was filtered to a lesser extent, retaining stereochemical information and reactions forming or breaking aromatic bonds. For one-step retrosynthesis, we evaluate on the USPTO\_full and USPTO\_50k datasets without reaction type, both of which have been used as benchmarks for retrosynthesis. We count a prediction as correct only if it matches the ground truth output SMILES exactly, including all stereochemistry but excluding atom mapping, after canonicalization by RDKit.

\subsection{Implementation details}
Our hyperparameters for D-GAT and D-GCN are adapted from GraphRetro \citep{Somnath_2020_GRAPHRETRO} and for the attention encoder from Molecular Transformer \citep{Schwaller_2019_MT} and GTA \citep{Seo_2021_GTA}. The number of message updating steps is set to $4$. Following Molecular Transformer and GTA, we fix the embedding and hidden sizes $d_{model}$ to $256$, the filter size for Transformer to $2048$, the number of attention heads to 8 for both D-GAT and Transformer, and the number of layers for the attention encoder and the Transformer decoder both to 6. We train our model using Adam optimizer \citep{Kingma_Ba_2015_Adam} with Noam learning rate scheduler \citep{Vaswani_2017_Transformer}. Similar to \citet{Schwaller_2019_MT}, we group reactions with similar number of SMILES tokens together, batch by the maximal number of token count, and scale the D-GAT outputs by $\sqrt{d_{model}}$ before feeding into the attention encoder. The details of the hyperparameters used for different datasets are summarized in Appendix \ref{appendix:param}. We save the model checkpoints every $5000$ steps, select the best checkpoints based on the top-1 accuracy on the validation sets, and report the performance on the held-out test sets. Beam search is used to generate the output SMILES during inference with a beam size of $30$. We filter out any SMILES that cannot be parsed by RDKit and keep the remaining as our final list of proposed candidates for evaluation. Our code for reproducing the results is available at \url{https://github.com/coleygroup/Graph2SMILES}.

\subsection{Results on reaction outcome prediction}
\label{subsection:forward}

Table \ref{table:forward} summarizes the results of Graph2SMILES and other existing works on reaction outcome prediction. We only include methods that perform evaluations on the USPTO\_480k\_mixed dataset in the table, and provide a brief comparison of other methods that only evaluate on the less challenging USPTO\_480k\_separated data in Appendix \ref{appendix:forward}. All excluded methods show inferior performance to the Molecular Transformer (MT), with the exception of NERF \citep{Bi_2021_NERF} which has a $0.3$ point improvement in top-1 accuracy.

\begin{table}[ht]
\caption{Results for reaction outcome prediction on USPTO\_480k\_mixed and USPTO\_STEREO\_mixed sorted by top-1 accuracy. Best results for complete columns are highlighted in \textbf{bold}.}
\label{table:forward}
\begin{center}
\renewcommand{\arraystretch}{1.0}
\begin{tabularx}{\textwidth}{l *{4}{>{\centering}X}}
\toprule
Methods  & \multicolumn{4}{c}{Top-$n$ accuracy (\%)} \tabularnewline
\cmidrule(lr){2-5}
&1  &3  &5  &10 \tabularnewline
\midrule
\multicolumn{5}{c}{\bf USPTO\_480k\_mixed} \tabularnewline
\midrule
MEGAN \citep{Sacha_2021_MEGAN}
&86.3   &92.4   &94.0   &95.4       \tabularnewline
Molecular Transformer \citep{Schwaller_2019_MT}
&88.6   &93.5   &94.2   &94.9       \tabularnewline
Graph2SMILES (D-GCN) (\textit{ours})
&90.3   &94.0   &94.6   &95.2       \tabularnewline
Graph2SMILES (D-GAT) (\textit{ours})
&90.3   &94.0   &94.8   &95.3       \tabularnewline
Augmented Transformer \citep{Tetko_2020_AT}
&90.6   &-      &\textbf{96.1}   &-          \tabularnewline
Chemformer \citep{Irwin_2021_Chemformer}
&\textbf{91.3}   &-      &93.7   &94.0       \tabularnewline
\midrule
\multicolumn{5}{c}{\bf USPTO\_STEREO\_mixed} \tabularnewline
\midrule
Molecular Transformer \citep{Schwaller_2019_MT}
&76.2   &84.3   &85.8   &-          \tabularnewline
Graph2SMILES (D-GAT) (\textit{ours})
&78.1   &84.5   &85.7   &86.7       \tabularnewline
Graph2SMILES (D-GCN) (\textit{ours})
&\textbf{78.1}   &\textbf{84.6}   &\textbf{85.8}   &86.8       \tabularnewline
\bottomrule
\end{tabularx}
\end{center}
\end{table}

For the USPTO\_480k\_mixed dataset, Graph2SMILES improves upon the MT baseline with $1.7$, $0.5$ and $0.4$ point increases in top-1, 3 and 10 accuracies respectively. Similarly, for the USPTO\_STEREO\_mixed dataset, Graph2SMILES improves the top-1 accuracy of the MT baseline by 1.9 points, with minor improvement for top-3 accuracy. While there is still a gap between Graph2SMILES and Augmented Transformer or Chemformer, our approach does not use test-time data augmentation and ensembling as in Augmented Transformer \citep{Tetko_2020_AT}, nor have we performed pretraining as in Chemformer \citep{Irwin_2021_Chemformer} whose models have up to 10 times as many parameters as Graph2SMILES. Ensembling and pretraining are potential directions for improving Graph2SMILES as they are still compatible with our backbone replacement for the Transformer. For reaction outcome prediction, there is a small advantage of using D-GAT over D-GCN, with improvements of up to 0.2 points on top-n accuracy.

\subsection{Results on one-step retrosynthesis}
\label{subsection:retrosynthesis}

We compare the results of one-step retrosynthesis on USPTO\_full of Graph2SMILES with all existing methods that report results on this dataset, to the best of our knowledge. As can be seen from Table \ref{table:USPTOfull}, Graph2SMILES achieves higher top-1 accuracy than all methods except GTA \citep{Seo_2021_GTA}, while not using any templates, atom mapping, or output-side data augmentation. These additional features or techniques are orthogonal to the graph-to-sequence architecture itself, and can therefore potentially improve Graph2SMILES. For example, atom mapping information can be used, either by injecting information about the reaction center as in RetroPrime \citep{Wang_2021_RetroPRIME}, or by augmentation with graph-truncated cross-attention as in GTA \citep{Seo_2021_GTA}.

The much smaller USPTO\_50k dataset has been benchmarked more extensively. We compare the results in Table \ref{table:USPTO50k}, marking only the usage of the same set of features and techniques used as in Table \ref{table:USPTOfull} for succinctness\footnote{Additional features and techniques used by other works include variational inference \citep{Kim_2021_Tied}, reranking \citep{Zheng_2020_SCROP,Sun_2020_EBM}, pretraining \citep{Irwin_2021_Chemformer} and dual loss \citep{Sun_2020_EBM}.}. The first group of rows shows that across methods that do not use reaction templates, atom mapping, or output SMILES augmentation, Graph2SMILES achieves the best top-1 accuracy, improving the Transformer baseline \citep{Lin_2020_AutoSynRoute} by $9.8$ points from $43.1$ to $52.9$. From our experiments, we observe that it is possible to boost top-n accuracies for $n$ > $1$ at the expense of sacrificing the top-1 accuracy (even slightly), creating some room for tradeoff. To avoid over-tuning and giving overly optimistic results, however, we only report the test results for models with the highest top-1 accuracy during validation. Unlike in reaction outcome prediction, using D-GAT does not have a clear advantage over D-GCN for retrosynthesis, yielding only improvement on top-5 and 10 accuracies for USPTO\_50k as in Table \ref{table:USPTO50k}.

\begin{table}[t]
\caption{Retrosynthesis results on USPTO\_full sorted by top-1 accuracy. \textit{Templ.}: reaction templates used; \textit{Map.}: atom-mapping required; \textit{Aug.}: output data augmentation used. Best results are highlighted in \textbf{bold}.  We exclude RetroXpert \citep{Yan_2020_RetroXpert} whose results for USPTO\_full have not been updated after the discovery of information leak.}
\label{table:USPTOfull}
\begin{center}
\renewcommand{\arraystretch}{1.0}
\begin{tabularx}{\textwidth}{lcc *{3}{>{\centering}X}}
\toprule
Methods  & \multicolumn{2}{c}{Top-$n$ accuracy (\%)} &
\multicolumn{3}{c}{Features / techniques used} \tabularnewline
\cmidrule(lr){2-3} \cmidrule(lr){4-6}
&1  &10 
&\small \textit{Templ.}  &\small \textit{Map.}
&\small \textit{Aug.} \tabularnewline
\midrule
RetroSim \citep{Coley_2017_RetroSim}
&32.8   &56.1   &\cmark &\cmark &\xmark \tabularnewline
MEGAN \citep{Sacha_2021_MEGAN}
&33.6   &63.9   &\xmark &\cmark &\xmark \tabularnewline
NeuralSym \citep{Segler_Waller_2017_NeuralSym}
&35.8   &60.8   &\cmark &\cmark &\xmark \tabularnewline
GLN \citep{Dai_2019_GLN}
&39.3   &63.7   &\cmark &\cmark &\xmark \tabularnewline
RetroPrime \citep{Wang_2021_RetroPRIME}
&44.1   &68.5   &\xmark &\cmark &\cmark \tabularnewline
Aug. Transformer \citep{Tetko_2020_AT}
&44.4   &\textbf{73.3}   &\xmark &\xmark &\cmark \tabularnewline
Graph2SMILES (D-GAT) (\textit{ours})
&45.7   &62.9   &\xmark &\xmark &\xmark \tabularnewline
Graph2SMILES (D-GCN) (\textit{ours})
&45.7   &63.4   &\xmark &\xmark &\xmark \tabularnewline
GTA \citep{Seo_2021_GTA}
&\textbf{46.6}   &70.4   &\xmark &\cmark &\cmark \tabularnewline
\bottomrule
\end{tabularx}
\end{center}
\end{table}

\begin{table}[ht]
\caption{Retrosynthesis results on USPTO\_50k without reaction type sorted by top-1 accuracy. \textit{Templ.}: reaction templates used; \textit{Map.}: atom-mapping required; \textit{Aug.}: output data augmentation used. Best results for each group of rows are highlighted in \textbf{bold}.}
\label{table:USPTO50k}
\begin{center}
\renewcommand{\arraystretch}{1.0}
\begin{tabularx}{\textwidth}{lcccc *{3}{>{\centering}X}}
\toprule
Methods  & \multicolumn{4}{c}{Top-$n$ accuracy (\%)} &
\multicolumn{3}{c}{Features / techniques used} \tabularnewline
\cmidrule(lr){2-5} \cmidrule(lr){6-8}
&1  &3  &5  &10 
&\small \textit{Templ.}  &\small \textit{Map.}    &\small \textit{Aug.}    \tabularnewline
\midrule
AutoSynRoute \citep{Lin_2020_AutoSynRoute}
&43.1   &64.6   &71.8   &\textbf{78.7}   &\xmark &\xmark &\xmark \tabularnewline
SCROP \citep{Zheng_2020_SCROP}
&43.7   &60.0   &65.2   &68.7   &\xmark &\xmark &\xmark \tabularnewline
GET \citep{Mao_2021_GET}
&44.9   &58.8   &62.4   &65.9   &\xmark &\xmark &\xmark \tabularnewline
Tied Transformer \citep{Kim_2021_Tied}
&47.1   &\textbf{67.2}   &\textbf{73.5}   &78.5   &\xmark &\xmark &\xmark \tabularnewline
Graph2SMILES (D-GAT) (\textit{ours})
&51.2   &66.3   &70.4   &73.9   &\xmark &\xmark &\xmark \tabularnewline
Graph2SMILES (D-GCN) (\textit{ours})
&\textbf{52.9}   &66.5   &70.0   &72.9   &\xmark &\xmark &\xmark \tabularnewline
\midrule
MEGAN \citep{Sacha_2021_MEGAN}
&48.1   &70.7   &78.4   &86.1   &\xmark &\cmark &\xmark \tabularnewline
G2Gs \citep{Shi_2020_G2Gs}
&48.9   &67.6   &72.5   &75.5   &\xmark &\cmark &\xmark \tabularnewline
RetroXpert \citep{Yan_2020_RetroXpert}
&50.4   &61.1   &62.3   &63.4   &\xmark &\cmark &\cmark \tabularnewline
GTA \citep{Seo_2021_GTA}
&51.1   &67.6   &74.8   &81.6   &\xmark &\cmark &\cmark \tabularnewline
RetroPrime \citep{Wang_2021_RetroPRIME}
&51.4   &70.8   &74.0   &76.1   &\xmark &\cmark &\cmark \tabularnewline
GLN \citep{Dai_2019_GLN}
&52.5   &69.0   &75.6   &83.7   &\cmark &\cmark &\xmark \tabularnewline
Aug. Transformer \citep{Tetko_2020_AT}
&53.2   &-      &80.5   &85.2   &\xmark &\xmark &\cmark \tabularnewline
LocalRetro \citep{Chen_Jung_2021_LocalRetro}
&53.4   &\textbf{77.5}   &\textbf{85.9}   &\textbf{92.4}   &\cmark &\cmark &\xmark \tabularnewline
GraphRetro \citep{Somnath_2020_GRAPHRETRO}
&53.7   &68.3   &72.2   &75.5   &\xmark &\cmark &\xmark \tabularnewline
Chemformer \citep{Irwin_2021_Chemformer}
&54.3   &-      &62.3   &63.0   &\xmark &\xmark &\cmark \tabularnewline
EBM (Dual-TB) \citep{Sun_2020_EBM}
&\textbf{55.2}   &74.6   &80.5   &86.9   &\cmark &\cmark &\cmark \tabularnewline
\bottomrule
\end{tabularx}
\end{center}
\end{table}

The second group of rows in Table \ref{table:USPTO50k} includes methods that use additional features or techniques. Graph2SMILES beats a number of methods in this group with a top-1 accuracy of $52.9$ without using any of templates, atom mapping or output SMILES augmentation. EBM (Dual-TB) \citep{Sun_2020_EBM} achieves the SOTA for top-1 accuracy, and LocalRetro \citep{Chen_Jung_2021_LocalRetro} for top-3, 5 and 10 accuracies respectively. Both make use of templates and atom mapping, which seem to be empirically helpful for this small dataset. Similar to how Graph2SMILES can benefit from using atom mapping as discussed earlier, we can make use of templates for potential gain, e.g. by retaining only candidates with reaction templates that have been seen in the training set.

\subsection{Ablation study}
\label{subsection:ablation}

We perform an ablation study to explore the effects of various components of Graph2SMILES by removing positional embedding or the global attention encoder. We summarize the results of using the D-GCN variant on USPTO\_50k, for which the quantitative effect is most conspicuous. Note that we have already shown the benefit of using graph representation over SMILES as well as using a D-MPNN encoder in Sections \ref{subsection:forward} and \ref{subsection:retrosynthesis}, where Graph2SMILES performs better than the Transformer baselines. From the results in Table \ref{table:ablation}, the removal of graph-aware positional embedding leads to a drop in top-1 accuracy of 2.1 points. The effect of removing the global attention encoder is more significant, decreasing the top-1 accuracy by 8.3 points. We therefore conclude that all three components of our encoder in Graph2SMILES, namely, the D-MPNN, the graph-aware positional embedding, and the global attention encoder, are important. They work collectively to capture a good representation of the input molecular graphs.
\begin{table}[ht]
\caption{Ablation study for Graph2SMILES (D-GCN) on USPTO\_50k.}
\label{table:ablation}
\begin{center}
\renewcommand{\arraystretch}{1.0}
\begin{tabularx}{\textwidth}{l *{4}{>{\centering}X}}
\toprule
Architecture     &Top-1  &Top-3  &Top-5  &Top-10 \tabularnewline
\midrule
Graph2SMILES (D-GCN)
&52.9   &66.5   &70.0   &72.9       \tabularnewline
\qquad no graph-aware positional embedding
&50.8   &65.2   &69.4   &73.6       \tabularnewline
\qquad no global attention encoder
&44.6   &60.7   &65.2   &69.6       \tabularnewline
\bottomrule
\end{tabularx}
\end{center}
\end{table}

\section{Discussion}
\label{section:discussion}
Throughout our experiments, to demonstrate the advantage over a vanilla Transformer, we have focused on the baseline Graph2SMILES model, forgoing the benefits of using additional features and techniques for performance engineering. We have already discussed how it is possible to integrate atom mapping or template-based filtering in Section \ref{subsection:retrosynthesis}. We briefly experiment with output-side SMILES augmentation, but the model as implemented becomes confused when trained to generate two equivalent but syntactically different SMILES. Techniques such as variational inference used in \citet{Chen_2020_Latent} and \citet{Kim_2021_Tied} may be required, which we leave as future work.

Although we have used the top-1 accuracy as a basis for comparison throughout our discussion, we recognize its limitations especially for retrosynthesis, which can have many equally plausible options. Similarly, the datasets we use for reaction outcome prediction are not perfectly detailed, with legitimate ambiguity in the identity of the major product. While Graph2SMILES shows strong performance for top-1 accuracy and can replace Transformer model with minimal modification to the pipeline, there could be cases when top-n accuracy is the more relevant metric (e.g. in some multi-step planning applications). In those cases, the aforementioned performance engineering techniques would be necessary to boost the top-n performance of Graph2SMILES.

\section{Conclusion}
In this paper, we present a novel Graph2SMILES model for template-free reaction outcome prediction and retrosynthesis. The permutation invariance of its D-MPNN and graph-aware positional embedding eliminates the need for any input-side SMILES augmentation, while achieving noticeable improvement over the Transformer baselines, especially for top-1 accuracy. Graph2SMILES is therefore an attractive drop-in replacement for any methods that use the Transformer model for molecular transformation tasks. Further gain may be possible through performance engineering tricks that are orthogonal to the architecture itself, which will be investigated in future work.

\newpage
\subsubsection*{Reproducibility Statement}
We hereby declare that all of our reported results are reproducible from our existing code base, subject to minor deviations due to hardware-level numerical uncertainties as we have observed for some GPU models (e.g. V100).

\subsubsection*{Acknowledgments}
This work was supported by the Machine Learning for Pharmaceutical Discovery and Synthesis consortium. The authors acknowledge the MIT SuperCloud and Lincoln Laboratory Supercomputing Center for providing HPC resources that have contributed to the research results reported within this paper. We thank Vignesh Somnath for assisting with D-MPNN. We also thank Samuel Goldman, Rocio Mercado, Thijs Stuyver, and John Bradshaw for commenting on the manuscript.

\bibliographystyle{iclr2022_conference}

\newpage
\appendix
\section{Appendix: atom and bond features used} \label{appendix:feature}
Table \ref{table:feature} summarizes the atom and bond features used in Graph2SMILES. Most features were adapted from GraphRetro \citep{Somnath_2020_GRAPHRETRO}, with the addition of chiral features (R/S and E/Z).

\begin{table}[ht]
\caption{Atom and bond features.}
\label{table:feature}
\begin{center}
\renewcommand{\arraystretch}{1.0}
\begin{tabularx}{\textwidth}{XXl}
\toprule
Feature    & Possible values   & Size  \\
\midrule
\multicolumn{3}{c}{\textbf{Atom Feature}}   \tabularnewline
\midrule
Atom symbol     & C, N, O etc.      & 65    \tabularnewline
Degree of the atom
& $\{d\in \gZ;0 \leq d \leq 9\}$     & 10    \tabularnewline
Formal charge of the atom
& $\{d\in \gZ;-2 \leq d \leq 2\}$   & 5   \tabularnewline
Valency of the atom
& $\{d\in \gZ;0 \leq d \leq 6\}$   & 7     \tabularnewline
Hybridization of the atom
& $sp$, $sp^2$, $sp^3$, $sp^3d$, $sp^3d^2$    & 5     \tabularnewline
Number of associated hydrogens
& 0, 1, 3, 4, 5         & 5     \tabularnewline
Chirality
& R, S, unspecified          & 3     \tabularnewline
Part of an aromatic ring
& True, false       & 2         \tabularnewline
\midrule
\multicolumn{3}{c}{\textbf{Bond Feature}}   \tabularnewline
\midrule
Bond type     & Single, double, triple, aromatic, other      & 5    \tabularnewline
Cis-trans isomerism
& E, Z, unspecified     & 3    \tabularnewline
Conjugated
& True, false     & 2    \tabularnewline
Part of a ring
& True, false   & 2   \tabularnewline
\bottomrule
\end{tabularx}
\end{center}
\end{table}

\section{Appendix: other methods for reaction outcome prediction}
\label{appendix:forward}

\begin{table}[ht]
\caption{Results for reaction outcome prediction on USPTO\_480k\_separated for methods excluded in Section \ref{subsection:forward}, sorted by top-1 accuracy. Best results are highlighted in \textbf{bold}.}
\label{table:forwardsep}
\begin{center}
\renewcommand{\arraystretch}{1.0}
\begin{tabularx}{\textwidth}{l *{3}{>{\centering}X}}
\toprule
Methods  & \multicolumn{3}{c}{Top-$n$ accuracy (\%)} \tabularnewline
\cmidrule(lr){2-4}
&1  &3  &5      \tabularnewline
\midrule
WLN/WLDN \citep{Jin_2017_WLDN}
&79.6   &87.7   &89.2       \tabularnewline
Seq2Seq \citep{Schwaller_2018_S2S}
&80.3   &86.2   &87.5       \tabularnewline
GTPN \citep{Do_2019_GTPN}
&83.2   &86.0   &86.5       \tabularnewline
WLDN5 \citep{Coley_2018_WLDN5}
&85.6   &92.8   &93.4       \tabularnewline
GRAT \citep{Yoo_2020_GRAT}
&88.3   &-      &-          \tabularnewline
Symbolic \citep{Qian_2020_Symbolic}
&90.4   &94.1   &95.0       \tabularnewline
NERF \citep{Bi_2021_NERF}
&\textbf{90.7}   &93.3   &93.7       \tabularnewline
\midrule
Molecular Transformer \citep{Schwaller_2019_MT}
&90.4   &\textbf{94.6}   &\textbf{95.3}       \tabularnewline
\bottomrule
\end{tabularx}
\end{center}
\end{table}

Table \ref{table:forwardsep} summarizes the results for methods excluded in Section \ref{subsection:forward} for reaction outcome prediction, most of which report the values on the less challenging USPTO\_480k\_separated dataset, in which the reagents have been heuristically separated from the reactants. Only NERF shows marginal improvement of 0.3 points for top-1 accuracy over Molecular Transformer, whereas all other methods cannot perform as well. We therefore use Molecular Transformer as our baseline in Section \ref{subsection:forward}. Note that ELECTRO \citep{Bradshaw_2018_ELECTRO} tests on a simpler subset of reactions with linear electron flow (LEF), and we therefore exclude it from the quantitative comparison.

\newpage
\section{Appendix: summary of four USPTO datasets used}
\label{appendix:dataset}

\begin{table}[ht]
\caption{Statistics of USPTO datasets used.}
\label{table:dataset}
\begin{center}
\renewcommand{\arraystretch}{1.0}
\begin{tabularx}{\textwidth}{ll *{3}{>{\centering}X}}
\toprule
Dataset     &Source     &Train size      &Validation size     &Test size       \tabularnewline
\midrule
USPTO\_480k\_mixed  &MT repo\dag
&409,035        &30,000     &40,000     \tabularnewline
USPTO\_STEREO\_mixed    &MT repo
&902,581        &50,131     &50,258     \tabularnewline
USPTO\_50k      &GLN repo\ddag
&40,008         &5,001      &5,007      \tabularnewline
USPTO\_full     &GLN repo
&810,496        &101,311    &101,311    \tabularnewline
\bottomrule
\multicolumn{5}{l}{\dag \small https://github.com/pschwllr/MolecularTransformer
\hfill \ddag \small https://github.com/Hanjun-Dai/GLN.}
\end{tabularx}
\end{center}
\end{table}

\section{Appendix: hyperparameter setting}
\label{appendix:param}

\begin{table}[ht]
\caption{Hyperparameter setting used in the experiments for different datasets. Best settings selected based on validation are highlighted in \textbf{bold} if multiple values have been experimented.}
\label{table:param}
\begin{center}
\renewcommand{\arraystretch}{1.0}
\begin{tabularx}{\textwidth}{Xll}
\toprule
Dataset        &Parameter      &Value(s) \tabularnewline
\midrule
All     &Embedding size     &\textbf{256}, 512        \tabularnewline
&Hidden size (same among all modules)    &\textbf{256}, 512       \tabularnewline
&Filter size in Transformer     &2048   \tabularnewline
&Number of D-MPNN layers     &2, \textbf{4}, 6        \tabularnewline
&Number of D-GAT attention heads     &8        \tabularnewline
&Attention encoder layers   &4, \textbf{6}        \tabularnewline
&Attention encoder heads   &8      \tabularnewline
&Decoder layers   &4, \textbf{6}        \tabularnewline
&Decoder heads   &8      \tabularnewline
&Number of accumulation steps   &4      \tabularnewline
\midrule
USPTO\_480k  &Batch type     &Source token counts    \tabularnewline
&Batch size     &4096   \tabularnewline
&Total number of steps  &300,000    \tabularnewline
&Noam learning rate factor  &2      \tabularnewline
&Dropout    &0.1        \tabularnewline
\midrule
USPTO\_STEREO  &Batch type     &Source token counts    \tabularnewline
&Batch size     &4096   \tabularnewline
&Total number of steps  &400,000    \tabularnewline
&Noam learning rate factor  &2      \tabularnewline
&Dropout    &0.1            \tabularnewline
\midrule
USPTO\_50k  &Batch type     &Source token counts    \tabularnewline
&Batch size     &4096   \tabularnewline
&Total number of steps  &200,000    \tabularnewline
&Noam learning rate factor  &2, \textbf{4}      \tabularnewline
&Dropout    &0.1, \textbf{0.3}            \tabularnewline
\midrule
USPTO\_full  &Batch type     &Source+target token counts    \tabularnewline
&Batch size     &8192   \tabularnewline
&Total number of steps  &400,000    \tabularnewline
&Noam learning rate factor  &2      \tabularnewline
&Dropout    &0.1            \tabularnewline
\bottomrule
\end{tabularx}
\end{center}
\end{table}

\end{document}